\documentclass[conference]{IEEEtran}
\IEEEoverridecommandlockouts
\usepackage{cite}
\usepackage{amsmath,amssymb,amsfonts}
\usepackage{algorithmic}
\usepackage{graphicx}
\usepackage{textcomp}
\usepackage{xcolor}
\def\BibTeX{{\rm B\kern-.05em{\sc i\kern-.025em b}\kern-.08em
    T\kern-.1667em\lower.7ex\hbox{E}\kern-.125emX}}

\begin{document}
 
\title{\vspace*{7.8mm} g2o vs. Ceres: Optimizing Scan Matching in Cartographer SLAM\\}
\author{\IEEEauthorblockN{1\textsuperscript{st} Quanjie Qiu}
\IEEEauthorblockA{ \textit{School of Engineering and Computer Science} \\
\textit{Laurentian University} \\
935 Ramsey Lake Rd.\\
Sudbury, ON, P3E 2C6, Canada. \\
quanjieqiu@gmail.com}
~\\
\and
\IEEEauthorblockN{2\textsuperscript{nd} MengCheng Lau}
\IEEEauthorblockA{\textit{School of Engineering and Computer Science} \\
\textit{Laurentian University} \\
935 Ramsey Lake Rd.\\
Sudbury, ON, P3E 2C6, Canada. \\
mclau@laurentian.ca}
*Corresponding author
~\\
}

\maketitle

\begin{abstract}
This article presents a comparative analysis of g2o and Ceres solvers in enhancing scan matching performance within the Cartographer framework. Cartographer, a widely-used library for Simultaneous Localization and Mapping (SLAM), relies on optimization algorithms to refine pose estimates and improve map accuracy. The research aims to evaluate the performance, efficiency, and accuracy of the g2o solver in comparison to the Ceres solver, which is the default in Cartographer. In our experiments comparing Ceres and g2o within Cartographer, Ceres outperformed g2o in terms of speed, convergence efficiency, and overall map clarity. Ceres required fewer iterations and less time to converge, producing more accurate and well-defined maps, especially in real-world mapping scenarios with the AgileX LIMO robot. However, g2o excelled in localized obstacle detection, highlighting its value in specific situations.
\end{abstract}


\section{Introduction}

Simultaneous Localization and Mapping (SLAM) is a fundamental challenge in robotics, where a robot must build a map of its environment while localizing itself within it. Cartographer, an open-source SLAM library developed by Google, supports real-time 2D and 3D mapping \cite{b1}. It integrates several key components—including sensor data processing, pose extrapolation, local and global SLAM modules—to perform accurate mapping and localization \cite{b2}. A core element in this pipeline is graph optimization, which estimates the system state by minimizing an error function defined over a graph.

In Cartographer's scan matching module, optimization solvers refine the robot's pose by minimizing the error between predicted and actual sensor measurements. The default solver, Ceres, is an open-source C++ library from Google designed to solve large-scale nonlinear least squares problems. It has been widely used in graph-based optimization frameworks \cite{b3, b4, b5}. Another well-known solver is g2o, an open-source framework tailored for graph-based nonlinear error minimization. g2o is commonly used in systems like ORB-SLAM \cite{b6, b7, b8} and SVO \cite{b9} for pose optimization and visual odometry, respectively.


While Ceres and g2o are prominent solvers, their comparative efficacy for scan matching within SLAM frameworks like Cartographer remains under-explored. This study addresses this gap, investigating their trade-offs to enhance overall SLAM efficiency, as scan matching is a key computational bottleneck. Identifying the better-suited solver is crucial for optimizing SLAM pipelines for faster, more robust real-time navigation amidst inherent computational demands. We compare Ceres and g2o, two widely-used SLAM solvers, to evaluate their effectiveness in Cartographer's scan matching. While other solvers like GTSAM exist, we limit our scope to these two to enable a focused and practical comparison with minimal architectural changes.

We chose the AgileX LIMO as our experimental platform, its suitability for autonomous navigation and SLAM research being clearly demonstrated by its adoption at Boston University (BU). BU's new RASTIC facility will simulate a Smart City, employing a fleet of 20-30 LIMOs for complex cooperative navigation tasks, underscoring the platform's capacity for demanding SLAM-dependent applications.

The following sections detail various aspects of the research. The background section provides a comprehensive literature review, offering insights into the theoretical frameworks of graph optimization and nonlinear least squares optimization, along with their applications in scan matching. The methodology section outlines the process of integrating g2o into Cartographer, detailing the necessary modifications and core components. The implementation section elaborates on the architectural design, the integration steps, and the rebuilding of Cartographer with g2o. Lastly, the experimental results and discussion sections present a performance comparison based on various metrics, followed by conclusions and recommendations for future work.

\section{Background}

The choice of optimization framework is critical in SLAM. g2o, introduced by Kümmerle et al. \cite{b10}, is a general graph optimization framework demonstrating performance comparable to specialized implementations, leveraging SLAM problem structures for efficient error minimization as detailed by Grisetti et al. \cite{b11}. Concurrently, the Ceres Solver, introduced by Agarwal et al. \cite{b12}, is another powerful C++ library for large-scale nonlinear least squares problems and serves as the default solver in Cartographer. The significance of solver selection is highlighted by comparative studies like Anela et al. \cite{b13}\cite{b14}, which evaluated frameworks including g2o and Ceres across various scenarios.

While optimizer performance can be context-dependent (e.g., Albert et al. \cite{b15} noted EKF outperforming g2o in specific AUV SLAM contexts), this paper focuses on the practical implications of substituting Ceres with g2o within Cartographer's scan matching module. The broader field of SLAM optimization also sees continuous advancements, such as tools for comparing nonlinear optimization techniques \cite{b16}, the integration of graph neural networks for pose graph optimization \cite{b17}, methods for graph sparsification \cite{b18}, and enhanced rotation estimation techniques \cite{b19}. This study aims to provide a focused comparison of g2o and Ceres directly within the Cartographer framework, contributing to a clearer understanding of their relative merits for this specific, widely-used SLAM system.

\section{Methodology}

In SLAM, pose inaccuracies can arise from sensor noise, odometry drift, and environmental complexity, significantly affecting localization and mapping precision \cite{b20}. One effective method to mitigate such errors is scan matching, which aligns incoming sensor data (e.g., laser scans or point clouds) with a map or previously acquired scans to refine the robot’s pose. By minimizing the discrepancy between current and reference data, scan matching enables precise real-time localization. Cartographer employs optimization algorithms, such as Ceres and g2o, to continuously update pose estimates and refine the map.

\subsection{Ceres Scan Matching}\label{AA}
\begin{figure}[htbp]
\centerline{\includegraphics[width=0.48\textwidth]{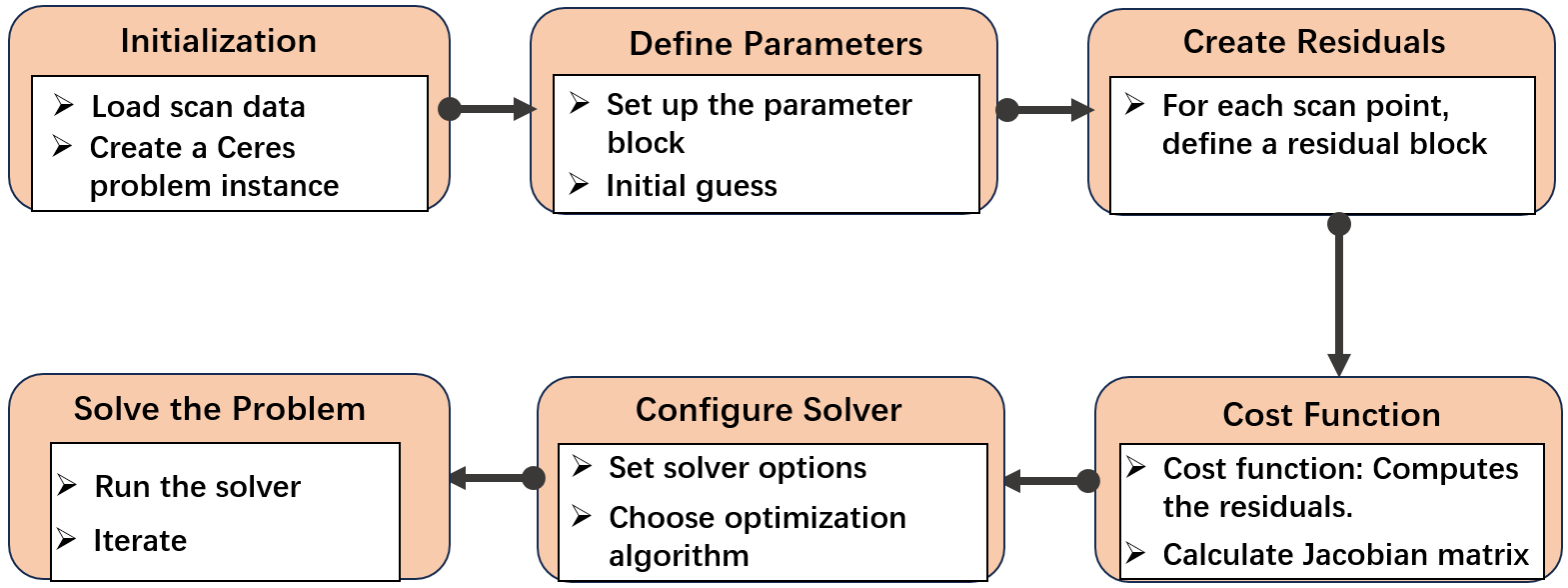}}
\caption{Flow diagram for Ceres handling scan matching.} 
\label{fig:ceres_fd} 
\end{figure}

Ceres Solver, Cartographer’s default optimizer, minimizes a least squares cost function during scan matching and is well-suited for large-scale nonlinear problems. Fig.~\ref{fig:ceres_fd} illustrates the process.

The key steps in Ceres include:
\begin{itemize}
\item Define Residuals and Cost Function: Residuals represent mismatches between transformed scan data and the map. The cost function sums squared residuals:
\begin{equation}
\cos t = \sum_i [r_i(x)]^2
\end{equation}
\item Setup Solver: Define residual blocks and the parameters to optimize.
\item Choose Algorithm: Select from Levenberg-Marquardt, Gauss-Newton, etc.
\item Numerical Techniques: Compute Jacobians and solve the system using Cholesky decomposition.
\item Convergence: Monitor and update parameters until convergence criteria are met.
\end{itemize}

\subsection{g2o} 


Replacing Ceres with g2o introduces a different backend for solving the nonlinear least squares problem of scan alignment. While both solvers are capable, they differ in structure, feature sets, and performance characteristics.

\begin{figure}[htbp]
\centerline{\includegraphics[width=0.48\textwidth]{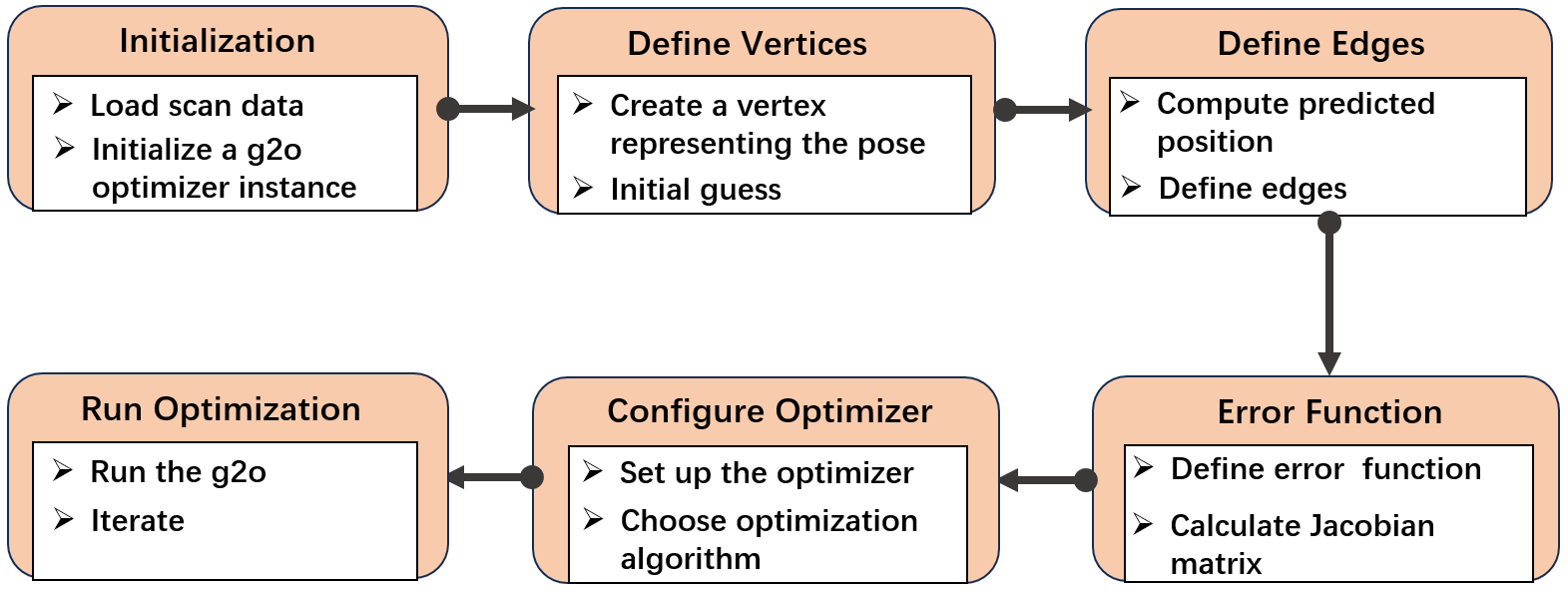}}
\caption{Flow diagram description for g2o handling scan matching.}
\label{fig:g2o_fd}
\end{figure}

In this study, the same scan matching formulation is retained, but g2o replaces Ceres to allow direct comparison of convergence, accuracy, and speed. Fig.~\ref{fig:g2o_fd} outlines the g2o-based workflow.

The key steps in g2o include:

\begin{itemize}
\item Initialization: Create data structures and set initial parameters for optimization.
\item Define Vertices: Represent robot poses or landmarks; each is a variable to optimize.
\item Define Edges: Represent constraints (e.g., odometry, loop closures) between poses with associated error functions.
\item Error and Jacobian: Compute errors between predicted and observed values; derive Jacobians to guide optimization.
\item Configure Optimizer: Choose a solver (e.g., Levenberg-Marquardt), and set convergence parameters.
\item Run Optimization: Iteratively adjust vertex estimates to minimize overall graph error until convergence.
\end{itemize}

\section{Implementation}

\subsection{Architecture}
\begin{figure}[htbp]
\centerline{\includegraphics[width=0.48\textwidth]{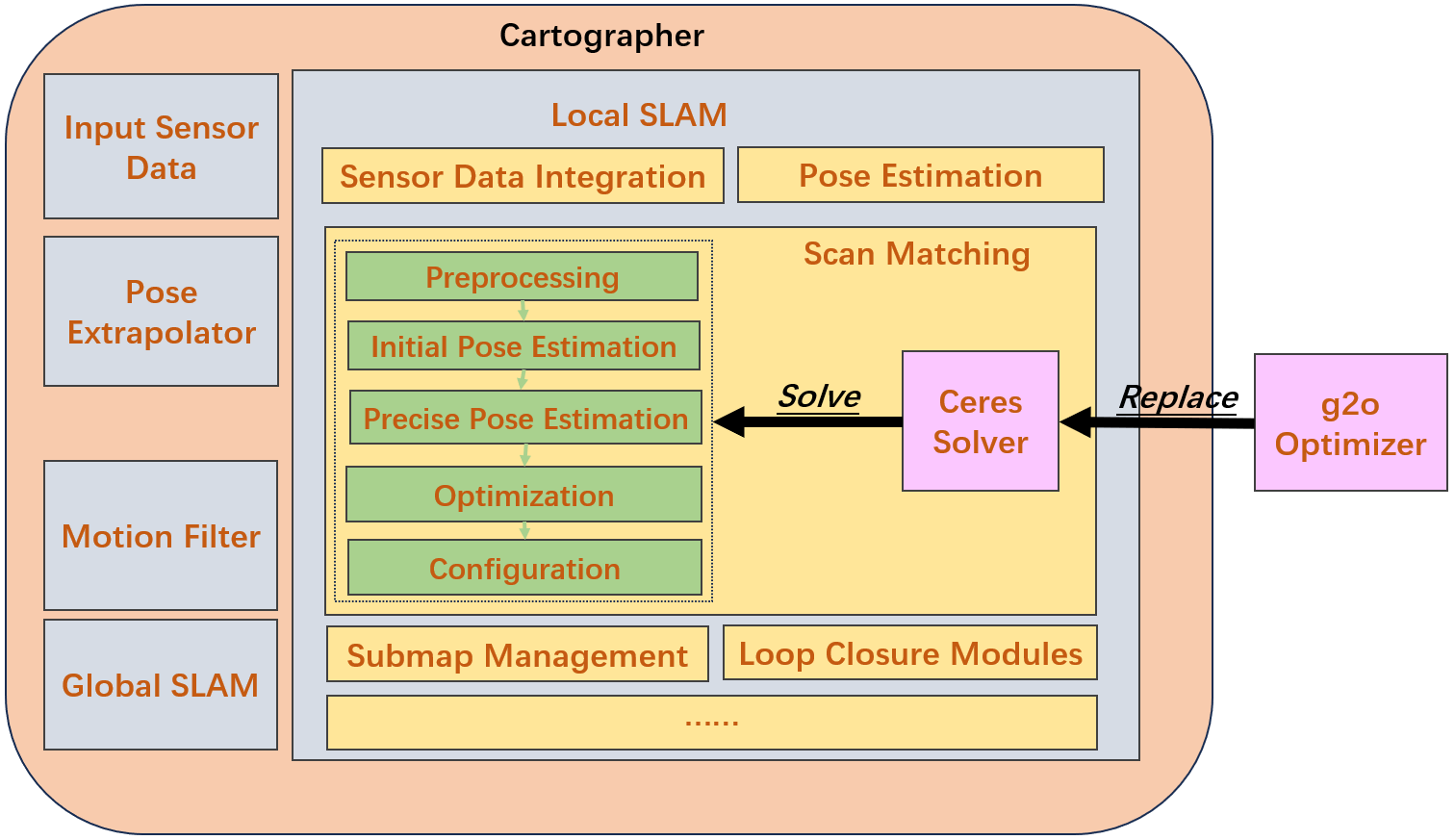}}
\caption{The architecture of scan matching. We use g2o optimizer to replace the Ceres Solver.}
\label{fig:architecture_scan_matching}
\end{figure}

Fig.~\ref{fig:architecture_scan_matching} outlines the architecture of Cartographer's local SLAM module, focusing on the scan matching process. In scan matching module, Ceres Solver is the default solver used in the precise pose estimation and optimization phases. Our main task is replacing the Ceres Solver with the g2o Optimizer for the optimization step in scan matching module of Cartographer's local SLAM.

\subsection{Defining Custom Data Type}
Ceres defines a template struct JetA using the Eigen library, performing various mathematical operations such as evaluation, differentiation, and integration. It is designed to represent a polynomial-like object for linear algebra operations. By using this data type, Cartographer can efficiently handle crucial mathematical operations in scan matching algorithms, which rely on optimizing transformations to align scans. What’s more, this can speed up the computations required during scan matching, such as transformations and distance calculations. It can also increase the customizability and flexibility while programming.

Since this data type is built into Ceres, we need to create a customized version. During the customization process, we can optimize and simplify the original data type to better suit our research needs.  The g2oJet struct is designed to hold a value and its derivatives. This is very useful in automatic differentiation, where we need to track both a function's value and its derivatives (we set it to 3 because in the scan matching component, we only use 3-dimensional calculations). The members is like:
\begin{itemize}
    \item {T a: Holds the value.}
    \item {Eigen::Matrix\texttt{<}T, 1, 3\texttt{>} v: Holds the derivatives as a 1x3 matrix. allowing for efficient computation of derivatives for functions involving multiple variables.}
\end{itemize}

The g2oJet supports a variety of operators and functions: arithmetic operations (addition, subtraction, multiplication, division), unary operations (negation), and common mathematical functions (exponential, logarithm, trigonometric functions). Overloads for operations with scalars and integration with Eigen's matrix operations are also provided, enabling seamless use in numerical computations and optimization tasks. This combination of constructors and operators makes JetA a versatile and powerful tool for differentiation and optimization applications.

\subsection{Defining g2o Vertex}
To meet the requirements of scan matching rephrase, we need to define custom vertices and edges which allow us to tailor the optimization process to the specific problems. Initially, we have a definition for a vertex class in a framework that deals with 3-dimensional vectors. A key component of this class is a method that applies an update to the current estimate of the vertex. This method takes a pointer to a double array, representing the update vector for the vertex's state. The raw update vector is mapped to a 3D vector type, providing a convenient way to handle the data as a 3D vector. Once the mapping is done, the method applies this update to the vertex's internal state.

The process of updating the estimate is straightforward yet essential. The current estimate is incremented by the mapped update vector. This operation is fundamental in optimization algorithms, where the vertex states are iteratively adjusted to minimize the overall error in the system. Each application of this update method refines the vertex's estimate, gradually converging toward a solution that best satisfies all the constraints in the system.

\begin{figure}[htbp]
\centerline{\includegraphics[width=0.48\textwidth]{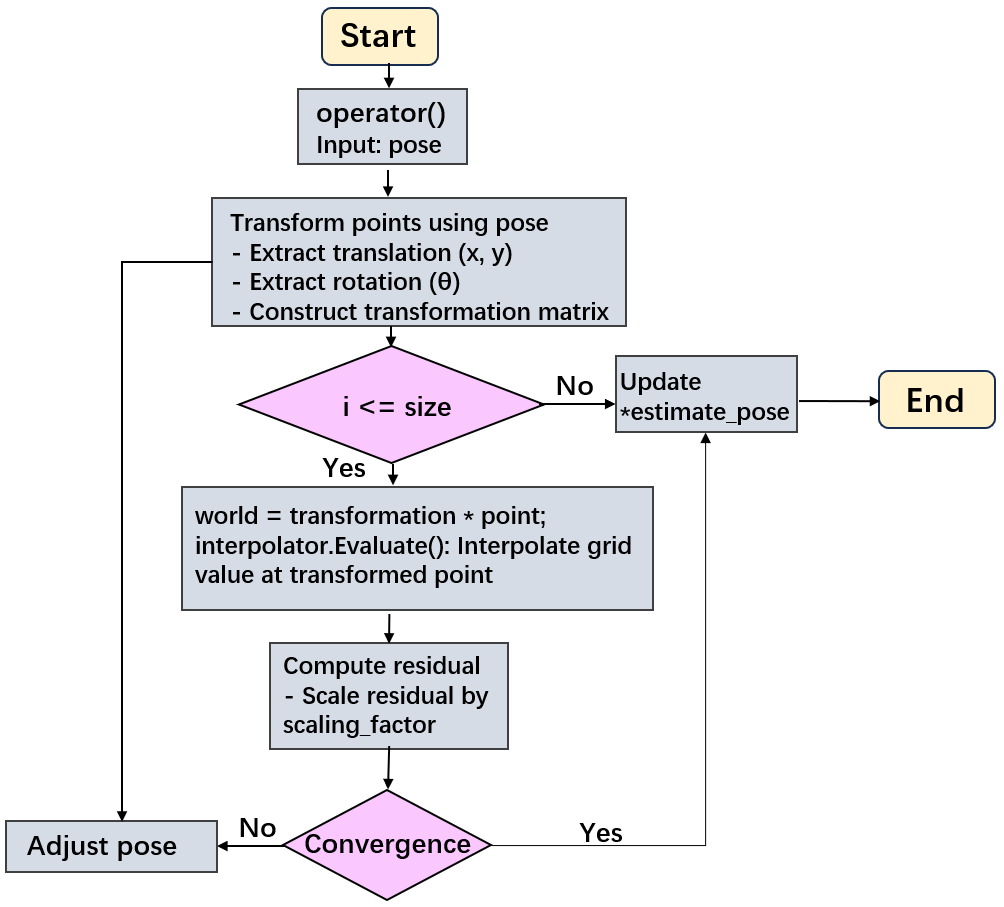}}
\caption{Process flow diagram of EdgeOccupiedSpace.}
\label{fig:edge_occupied_space}
\end{figure}

\subsection{Defining g2o Edges}
Three constraints are most important for optimizing vertex, they are:
\begin{itemize}
    \item EdgeOccupiedSpace: This constraint creates a cost function that penalizes discrepancies between the occupied spaces in the map and the occupied spaces in the new scan. It ensures that the robot’s pose aligns with the environment as represented in the map. Fig.~\ref{fig:edge_occupied_space} illustrates the steps of the EdgeOccupiedSpace class and its operator() function for computing the cost of matching a point cloud to a grid map given a pose.

\item EdgeTranslationDelta: This constrains creates a cost term that penalizes deviations in the translation component (the robot's x and y coordinates) between poses. It helps in aligning the new pose with the expected or previous poses by minimizing translation errors.
\item EdgeRotationDelta: This constraint creates a cost term that penalizes deviations in the rotation component (the robot’s orientation) between poses. It ensures that the robot’s orientation is consistent with the expected or previous orientations.
\end{itemize}

\subsection{Scan Matching using g2o}
\begin{figure}[htbp]
\centerline{\includegraphics[width=0.48\textwidth]{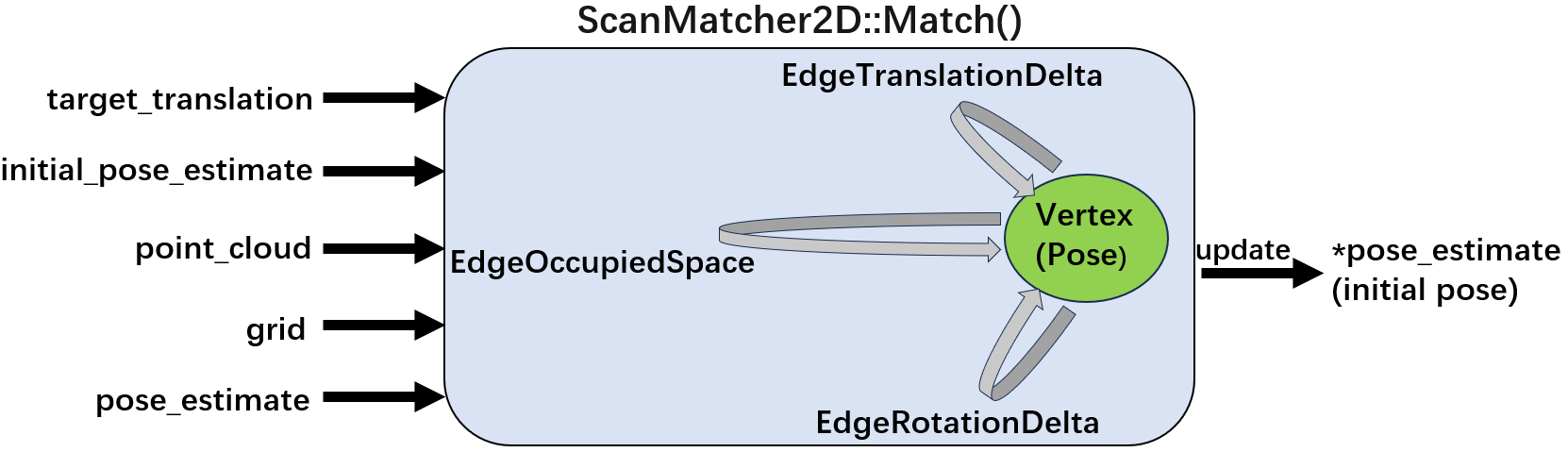}}
\caption{Working principle of ScanMatcher2D::Match().}
\label{ScanMatcher2D_Match()}
\end{figure}

The function ScanMatcher2D::Match()
is the key component of scan matching because it encapsulates this entire process of error calculation and pose adjustment. This function plays a crucial role in aligning scans to submaps, ensuring accurate mapping and localization. Fig.~\ref{ScanMatcher2D_Match()} shows the working principle.

\begin{itemize}
    \item Input: The target\_translation is the desired target translation for the scan. The initial\_pose\_estimate is an initial estimate of the robot's pose, which serves as the starting point for optimization. The point\_cloud is the current scan data represented as a point cloud. The grid is the submap or occupancy grid to which the scan is being matched. The pose\_estimate is the current pose estimate that will be updated.
    \item Vertex (Pose): The central element in the optimization process, representing the pose of the robot. It gets updated iteratively based on the optimization criteria defined by the edges.
    \item Edges (Constraints): EdgeOccupiedSpace evaluates the alignment of the point cloud with the occupied space in the grid. EdgeTranslationDelta assesses the difference in translation between the current and target translations. EdgeRotationDelta evaluates the difference in rotation between the current and target orientations.
    \item Output: The pose\_estimate is updated with the optimized pose, providing a refined estimate of the robot's position and orientation.
\end{itemize}

\begin{figure}[htbp]
\centerline{\includegraphics[width=0.4\textwidth]{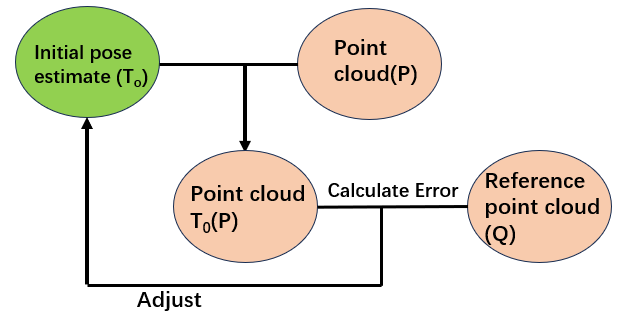}}
\caption{The error calculation and adjustment steps.}
\label{fig:steps}
\end{figure}

The process begins with an initial pose estimate, denoted as \(T_{\text{0}}\). This is the initial guess of the robot's pose before any scan matching is applied. The Cloud Point (P) represents the individual points within the point cloud obtained from the current scan. It is going to be transformed using the initial pose estimate, resulting in \(T_{\text{0}}\)(P). This step aligns the scan data based on the initial pose. Fig.~\ref{fig:steps} breaks down the scan matching process by detailing the error calculation and adjustment steps.

The reference point cloud, Q, represents the submap or grid data that the current scan is being matched against. This is the known map data used for comparison. The transformed point cloud \(T_{\text{0}}\)(P) is compared with the reference point cloud Q. The error between these two sets of points is calculated. This error quantifies the misalignment between the current scan and the submap. Based on the calculated error, adjustments are made to the initial pose estimate \(T_{\text{0}}\). The pose is iteratively updated to minimize the error, resulting in a refined pose estimate that better aligns the point cloud with the reference point cloud.

\subsection{Rebuild Cartographer with g2o}
To integrate the g2o optimization framework into Cartographer's existing infrastructure to handle pose graph optimization, we have prepared the modified g2o's code, written the CMake files, and ensured that g2o and all its dependencies are correctly located and linked within the CMake configuration. After compilation, we now have two versions of Cartographer available in the environment: the original one and the modified version where the Ceres computing library's scan matching part has been replaced with g2o.

\section{Experiments}
\subsection{Experiments on randomly generated point clouds}

The experiment was conducted on an ASUS TUF Gaming F15 FX507VV\_FX507VV laptop, equipped with an Intel Core i9-13900 (20 CPUs) operating at 2.60GHz, and 16GB of RAM. The system operated on Ubuntu 22.04 with ROS 2 Humble installed. The tolerance for residual error reduction was set to $10^{-10}$.

The experiment involved a loop of 20 iterations, where each iteration represented a distinct run using randomly generated point clouds (consisting of 5 points) and an initial pose. Each point in the cloud comprised an x-coordinate, a y-coordinate, and an orientation $\theta$. In each iteration, the point cloud and initial pose were fed into two versions of Cartographer: one using Ceres and the other using g2o for scan matching. Both Cartographers utilized their respective graph optimization frameworks to refine the initial pose by aligning it with the point cloud and the probability grid. This setup enabled a direct comparison of the performance of Ceres and g2o in optimizing the initial pose.

\begin{figure}[htbp]
\centerline{\includegraphics[width=0.48\textwidth]{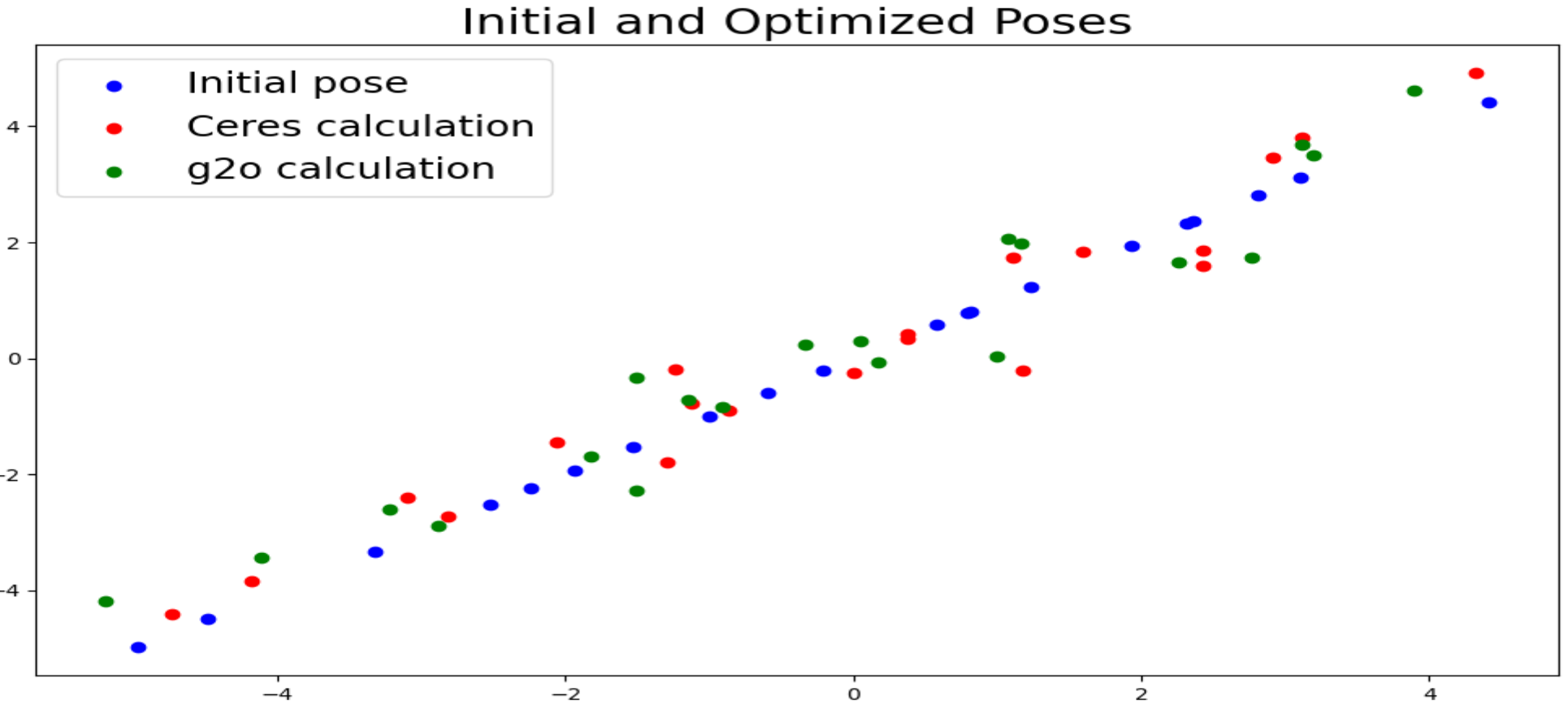}}
\caption{The calculation results by Ceres and g2o.}
\label{fig:points}
\end{figure}

The experiment result is shown in Fig.~\ref{fig:points}. To better illustrate the outcome, the randomly generated initial points (blue) were arranged in a straight line. After optimization, Ceres-adjusted poses are shown in red, and g2o-adjusted poses in green. Both solvers successfully adjusted the poses using automatic differentiation, and their results were generally close, indicating effectiveness in pose optimization.

\begin{table}[htbp]
\centering
\caption{Initial Pose Optimization: Ceres vs. g2o Comparison}
\label{tbl:RMSE}
\begin{tabular}{|l|c|}
\hline
\textbf{Solver} & \textbf{RMSE} \\
\hline
Ceres &  0.49624 \\
g2o & 0.51640 \\
\hline
\end{tabular}
\end{table}

However, a few discrepancies between the two were noted, attributed to differences in implementation details, cost functions, and parameter tuning. Quantitatively, Ceres achieved a slightly lower RMSE of 0.49624 compared to 0.51640 for g2o as shown in Table~\ref{tbl:RMSE}. 

\begin{table}[htbp]
\centering
\caption{Comparison of Optimization Accuracy and Time Consumption}
\label{tbl:accuracy_time}
\begin{tabular}{|l|c|c|}
\hline
\textbf{Solver} & \textbf{Time (ms)} & \textbf{Average Iteration} \\
\hline
Ceres & \centering 1.14 &  13 \\
g2o & \centering 1.59 &  23 \\
\hline
\end{tabular}
\end{table}

In terms of efficiency, Ceres was also faster, requiring only 1.14 milliseconds and an average of 13 iterations to converge, whereas g2o needed 1.59 milliseconds and 23 iterations on average as shown in Table~\ref{tbl:accuracy_time}. These findings suggest that while both solvers are capable, Ceres demonstrates superior convergence speed and computational efficiency in this experimental setup.




\subsection{Experiments with a Real Robot}

For this experiment, we used an AgileX LIMO ROS2 equipped with a NUC, EAI T-mini Pro Lidar, and depth cameras to evaluate the system's performance in a controlled, real-world-like environment. The experiment took place in an indoor lab space designed to simulate a variety of real-world conditions. The environment was populated with diverse features and obstacles, including walls, furniture, desks, and boxes, providing a complex and dynamic setting for testing the robot's navigation and mapping capabilities.

The goal of the experiment was to assess the robot's ability to accurately adapt to different environmental conditions and validate the effectiveness of its onboard algorithms, including SLAM and scan matching. The presence of various obstacles and features challenged the robot’s sensors and optimization frameworks, offering a robust testbed for measuring how well the robot could handle localization and mapping tasks using different optimization methods, such as Ceres and g2o.

\begin{figure}[htbp]
  \centering
  \includegraphics[width=0.48\textwidth]{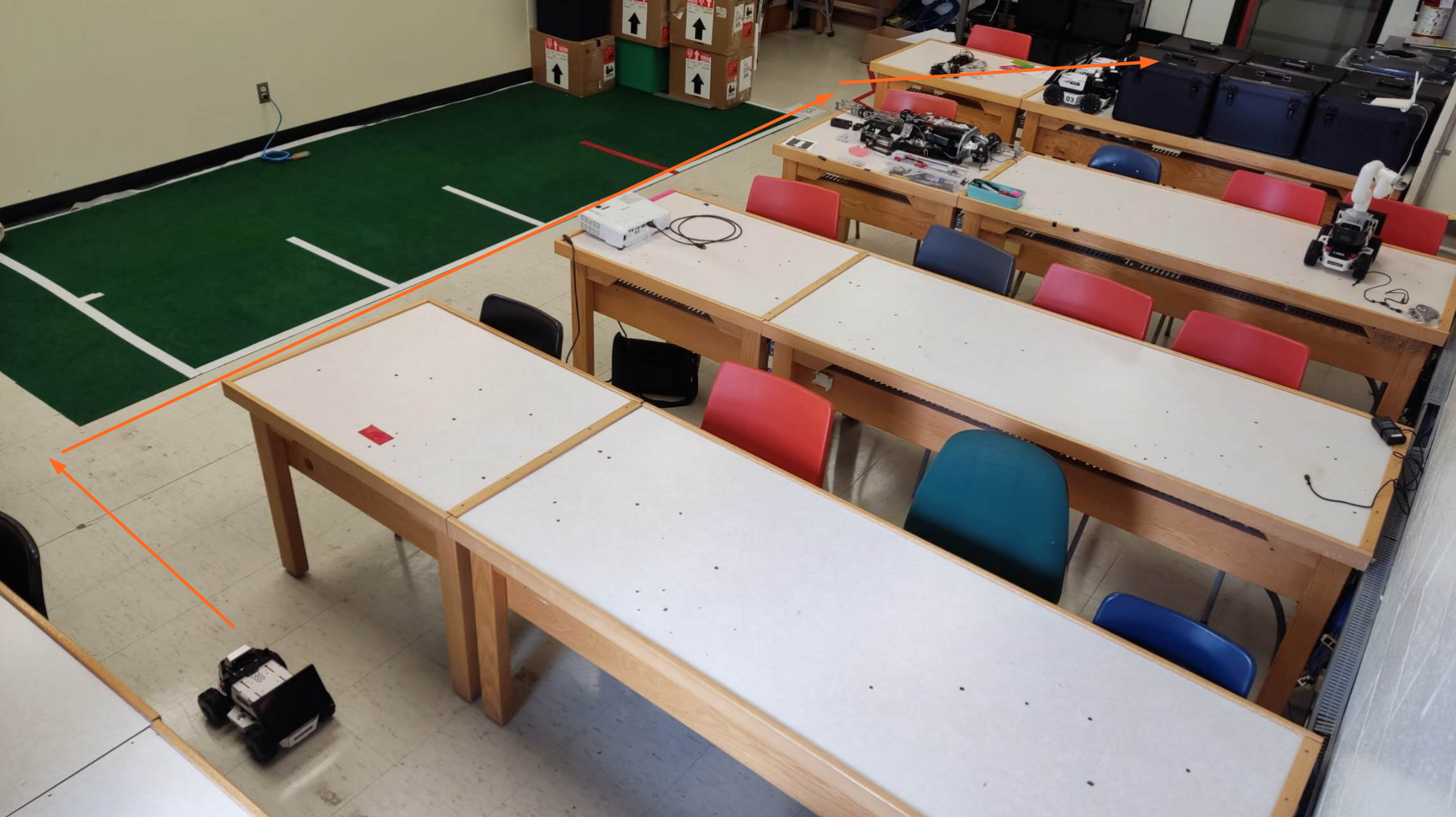}
  \caption{The lab environment setup and the fixed movement path (orange arrows) of the robot during the experiment.}
  \label{fig:lab}
\end{figure}

\subsection{Fixed Movement Path}

To evaluate the performance of different algorithms, we deployed two instances of Cartographer on the robot, allowing it to switch between the two algorithm packages during indoor map building. The robot was manually driven along a predefined path in the lab at a constant speed to ensure consistency in the experiment. The lab environment is illustrated in Fig.~\ref{fig:lab}, where the robot's driving path is marked by the orange arrow. This path was carefully selected to maximize the lidar’s coverage of the lab space, ensuring comprehensive perception of the environment.

\begin{figure}[htbp]
  \centering
  \includegraphics[width=0.48\textwidth]{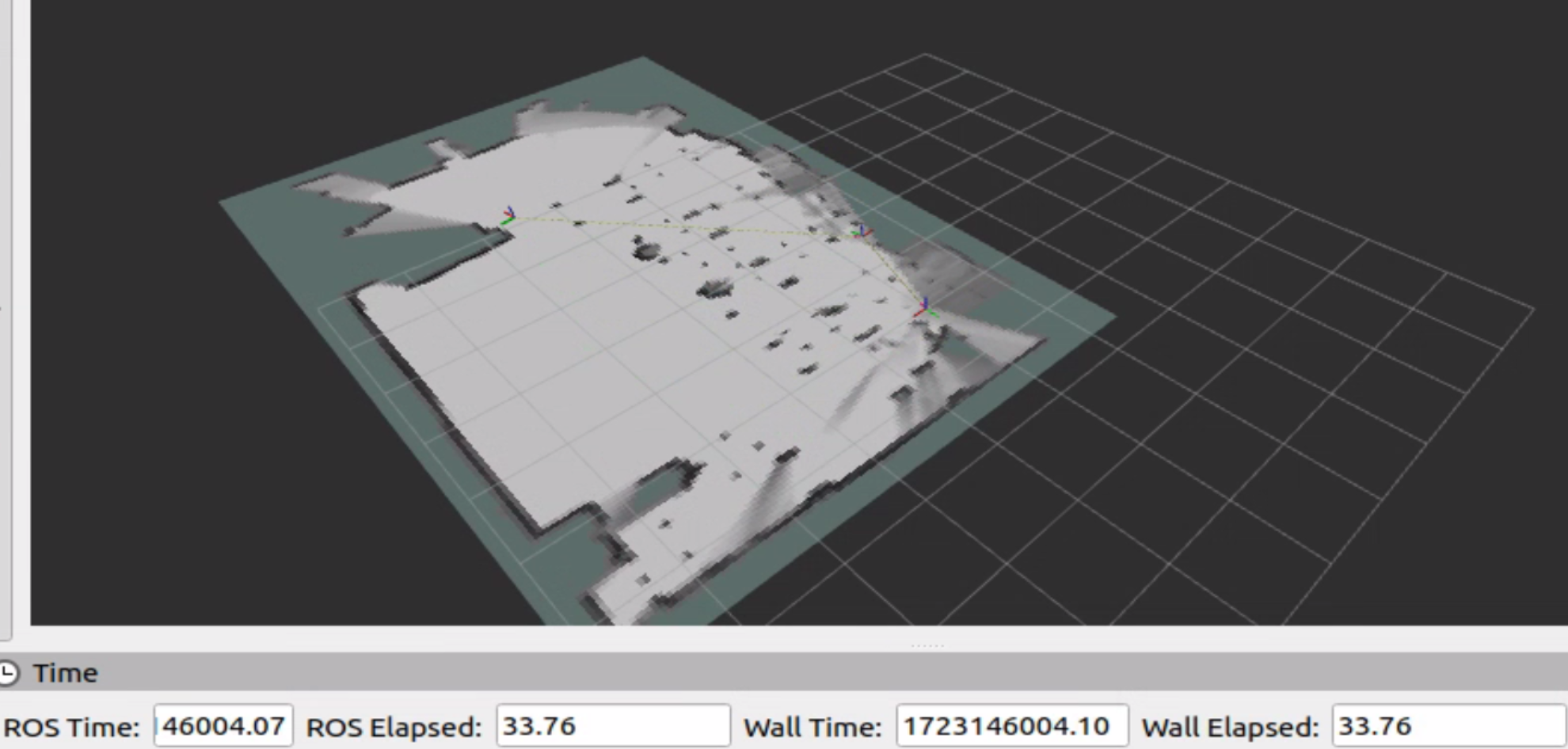}
  \caption{Map building with Cartographer (Ceres) in a limited time.}
  \label{fig:time_Ceres}
\end{figure}

\begin{figure}[htbp]
  \centering
  \includegraphics[width=0.48\textwidth]{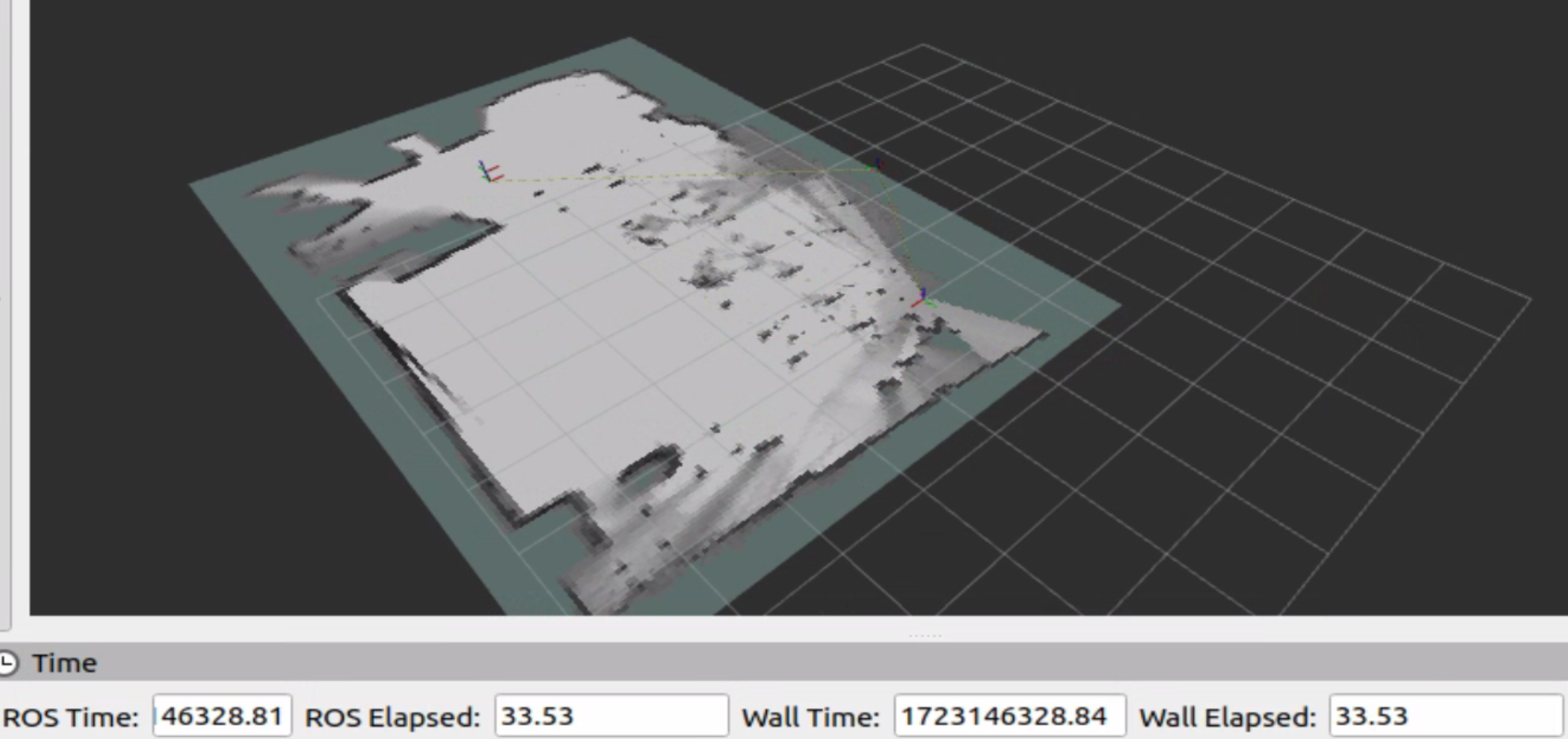}
  \caption{Map building with Cartographer (g2o) in a limited time.}
  \label{fig:time_g2o}
\end{figure}

With the robot followed the same path, the driving time for both setups set to approximately 30 seconds. This setup allowed us to assess and compare how quickly each version of Cartographer (using Ceres and g2o) could converge within the limited time frame. The resulting maps are displayed in Fig.~\ref{fig:time_Ceres} and Fig.~\ref{fig:time_g2o}.


In comparison, Cartographer using Ceres generally produces better results than the g2o implementation, particularly in terms of clearer wall outlines and more accurate obstacle distribution. For instance, at key locations like the endpoint (e.g., the lower left corner of the room), the g2o version exhibits weaker map closure, indicating slower overall map-building performance. Over the same time period, the Ceres version completes most of the structural models, whereas the g2o version struggles to perform loop closure in many areas. This highlights the slower convergence speed of g2o, as observed in the simulated point cloud experiments. Nevertheless, the g2o version does outperform Ceres in certain localized regions, such as along the wall in the top right corner of the room.

\section{Conclusions}

This study compared Ceres and g2o solvers within Cartographer’s scan matching module through synthetic and real-world experiments. Ceres consistently outperformed g2o in terms of speed and convergence, requiring fewer iterations and less time while achieving slightly better accuracy. In real-world tests with the AgileX LIMO robot, Ceres produced clearer and more complete maps, while g2o lagged in map closure and speed but showed strengths in localized obstacle detection. These results highlight Ceres’s overall efficiency and suitability for general SLAM tasks, whereas g2o may be advantageous in specific, localized scenarios.

It should be noted that Ceres benefits from native integration and extensive tuning within Cartographer, unlike our custom g2o implementation. Additionally, manual control during testing may have introduced minor variability, though repeated trials were conducted to ensure result consistency. Overall, this work offers practical insight into solver selection for SLAM systems and underscores the trade-offs between speed, accuracy, and adaptability when integrating alternative optimization backends.

Our future work includes further tuning of the g2o implementation, integrating other solvers such as GTSAM for broader comparison, and evaluating performance in more complex or dynamic environments. Additionally, automated benchmarking and real-time deployment on multi-robot systems could further validate the scalability and robustness of alternative solvers in SLAM pipelines.

\vspace{12pt}

\end{document}